\documentclass[10pt,twocolumn,letterpaper]{article}

\usepackage[pagenumbers]{cvpr}

\usepackage{pifont}

\makeatletter
\patchcmd{\@maketitle}{\vskip .375in}{\vskip 3pt}{}{}
\patchcmd{\@maketitle}{\vspace*{24pt}}{\vspace*{6pt}}{}{}
\makeatother

\definecolor{cvprblue}{rgb}{0.21,0.49,0.74}
\usepackage[pdfversion=1.7,breaklinks,colorlinks,allcolors=cvprblue]{hyperref}

\def\confName{CVPR}
\def\confYear{2027}

\title{BFA-WAM: Bridging Future Motion and Robot Action Generation}

\author{%
  Fengyuan Liu\textsuperscript{1}\quad
  Fu Rong\textsuperscript{1}\quad
  Shaoyu Chen\textsuperscript{2}\quad
  Hui Sun\textsuperscript{2}\quad
  Qian Zhang\textsuperscript{2}\quad
  Lefei Zhang\textsuperscript{1}\thanks{Corresponding author.}\\
  {\normalsize\textsuperscript{1}\,School of Computer Science, Wuhan University}\\
  {\normalsize\textsuperscript{2}\,Horizon Robotics}%
}
\hypersetup{pdfauthor={Fengyuan Liu, Fu Rong, Shaoyu Chen, Hui Sun, Qian Zhang, Lefei Zhang}}

\begin{document}
\maketitle
\begin{abstract}
World action models (WAMs) use predictive world representations to guide robotic action generation.
However, appearance-oriented prediction supervises motion only indirectly, while token-wise attention does not explicitly turn world information into shared modulation of action-layer updates.
We introduce BFA-WAM, a framework that combines explicit future-motion supervision with world-conditioned action refinement.
RGB--flow fusion (RFF) allows action queries to attend jointly to current RGB features and flow-supervised future features, linking anticipated displacement to visible objects and their spatial context.
The world model jointly learns future RGB and optical-flow prediction, with an additional one-step flow reconstruction objective supervising future-motion features under the input conditions used at deployment.
Complementing token-wise feature retrieval, world-conditioned modulation (WCM) uses separate observation and motion branches to summarize observed RGB and future-flow features.
These summaries produce layer-wise shifts, scales, and gates shared across action positions, conditioning action refinement on the observed scene and anticipated motion.
At inference, a single world-model pass supplies both the attention cache and modulation residuals, which are reused throughout action denoising without decoding future frames or estimating optical flow online.
Extensive experiments across multiple benchmarks demonstrate the strong performance of BFA-WAM.
Code is available at \mbox{\url{https://bfa-wam.github.io/}}.\par\vspace{-\baselineskip}
\end{abstract}
  
\section{Introduction}
\label{sec:intro}

Developing general-purpose robots requires policies that can interpret their surroundings and anticipate the changes caused by physical interaction.
World action models (WAMs) pursue this goal by learning future visual prediction alongside action generation~\cite{yuan2026fastwam}.
While standard vision-language-action (VLA) models learn to map observations and language instructions to robot actions, WAMs additionally use the evolution of visual states as a learning signal.
Predicting these changes can help the model capture object motion and interaction dynamics, providing world representations that inform action generation.

\begin{figure*}[t]
    \centering
    \includegraphics[width=\textwidth]{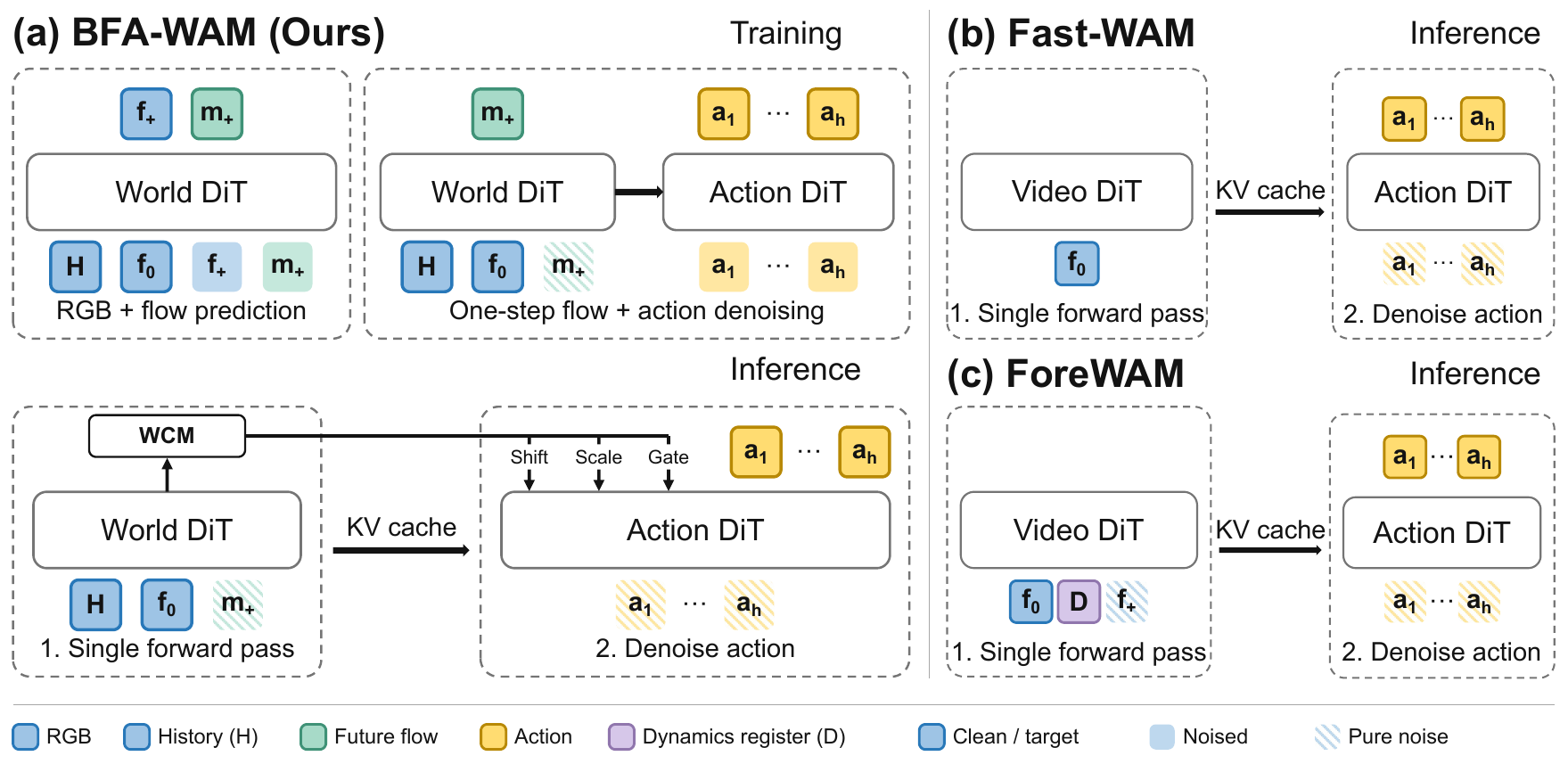}
    \caption{\textbf{What the world model supplies for action denoising.}
    All three policies run a world diffusion transformer (DiT) once per decision and reuse its features during action denoising.
    (a) BFA-WAM learns future motion from RGB and flow supervision (top). At inference (bottom), current RGB and future-flow features supply the attention cache, and the same pass provides world-conditioned modulation (WCM) to the action layers.
    (b) Fast-WAM caches features of the current observation. (c) ForeWAM adds future slots and dynamics registers to the cache.}
    \label{fig:wam-comparison}
\end{figure*}

The value of world modeling for manipulation depends on both what its representations encode and how they guide action generation.
Effective guidance should make anticipated motion explicit and allow world information to shape the refinement of the action sequence as a whole.
Existing architectures face two challenges in achieving these goals.

\textbf{First}, appearance-oriented world information provides only indirect cues about motion.
As shown in \cref{fig:wam-comparison}(b), Fast-WAM~\cite{yuan2026fastwam} retains video prediction during training and uses a single world-model pass to encode the current observation for action generation.
ForeWAM~\cite{huang2026forewam} further introduces future keys and values (Future-KV), giving the action expert access to predictive features after a single world-model pass, as shown in \cref{fig:wam-comparison}(c).
However, appearance prediction captures motion implicitly through visual changes, alongside information about object color and texture.
Optical flow provides a direct representation of motion by explicitly encoding the direction and magnitude of displacement across image regions.
This motivates using future optical flow as a direct training target for the features supplied to the action expert, providing motion cues that complement the appearance information in current RGB observations.
\textbf{Second}, these architectures lack an explicit mechanism that turns world information into a shared global condition for action-layer updates.
Cross-attention to language tokens provides fine-grained semantic guidance, while attention to world features supplies visual context for individual action tokens.
Modulation offers a complementary form of guidance by using a shared global semantic condition to adapt how the action network processes its internal features.
Conditioning layer-wise modulation on the observed scene and anticipated motion can therefore provide global guidance throughout action refinement.

To address these challenges, we introduce BFA-WAM, a world action framework for language-conditioned robotic manipulation.
As shown in \cref{fig:wam-comparison}(a), our core idea is to fuse current RGB appearance with anticipated motion represented by optical flow to guide action generation.
Specifically, we devise \textbf{RGB--flow fusion (RFF)}, in which current RGB features describe visible objects and their spatial context, while flow-supervised future features provide cues about their anticipated motion.
These features are supplied as current RGB keys and values (K/V) and flow-based Future-KV, allowing action queries to fuse appearance and motion through joint attention.
By combining these complementary cues, RFF enables the action expert to relate anticipated displacement to the objects and regions visible in the current scene.

To further condition action-layer updates on the observed scene and anticipated motion, we design \textbf{world-conditioned modulation (WCM)}.
Here, the conditioning information comes from the same world-model pass that supplies RFF with RGB and future-flow features.
WCM consists of an observation branch and a motion branch.
The observation branch pools current and historical RGB features and maps their summaries to layer-wise modulation residuals, conditioning action refinement on the observed scene.
Similarly, the motion branch summarizes future-flow features to provide additional modulation residuals that condition action updates on anticipated motion.
Together, the two branches adjust the shifts, scales, and gates of the attention and feed-forward sublayers, with the modulation shared across action positions.

Experimental results on multiple robotic manipulation benchmarks, including LIBERO-Plus, demonstrate the strong performance of BFA-WAM.
Our main contributions are summarized as follows:
\begin{itemize}
    \item We propose BFA-WAM, a world action model that integrates RGB--flow fusion and world-conditioned modulation to guide robotic action generation. At inference, a single world-model pass provides both forms of guidance, which are reused throughout action denoising without explicit future-video generation.
    \item We devise RFF, which fuses current RGB appearance with flow-supervised future motion through joint attention, providing complementary guidance to the action expert via current-observation K/V and Future-KV.
    \item We develop WCM, which uses summaries of the observed scene and anticipated motion to modulate the action layers. By adjusting the shifts, scales, and gates of attention and feed-forward updates, WCM complements token-wise feature retrieval with shared conditioning across action positions.
\end{itemize}

\section{Related Work}
\label{sec:related_work}

\paragraph{Vision-Language-Action Policies.}
VLA models transfer pretrained vision-language representations to robot control, as in RT-2~\cite{zitkovich2023rt2} and OpenVLA~\cite{kim2024openvla}.
Diffusion and flow-matching policies generate continuous action chunks~\cite{chi2025diffusionpolicy,black2025pi0,octo2024,liu2025rdt}, while diverse robot pretraining and heterogeneous co-training improve generalization~\cite{bjorck2025gr00t,black2025pi05,openx2024}.
Action prediction alone does not explicitly supervise scene evolution during execution, motivating world modeling for policy learning.

\paragraph{World Action Models.}
World action models (WAMs) couple predictions of how a scene evolves with the generation of robot actions.
Cascaded WAMs use predicted video plans~\cite{du2023unipi}, geometric motion representations~\cite{xu2024im2flow2act,zhi2025flowaction}, or predictive visual features~\cite{hu2024vpp} to guide a downstream controller in generating actions.
In policy mode, FlowWAM~\cite{chen2026flowwam} jointly denoises future RGB and optical-flow latents, then decodes robot actions from the generator's intermediate features.
Joint WAMs model future observations and actions together. Autoregressive models arrange visual and action tokens in a common sequence~\cite{cheang2024gr2,wu2024gr1}, while diffusion and flow models use a shared backbone~\cite{zhu2025uwm,shen2025videovla} or connect visual and action streams through attention~\cite{yang2025covar}, intermediate features~\cite{ma2026dit4dit}, or shared representations~\cite{li2025uva}.
Repeatedly updating future visual states during action generation can increase inference cost. Fast-WAM~\cite{yuan2026fastwam} and GigaWorld-Policy~\cite{ye2026gigaworld} use world prediction as a training objective and generate actions directly at deployment.
Other methods give the policy access to predictive features more directly~\cite{tian2025seer}. Enfold~\cite{zeng2026enfold} distills features from a world generator, and Robust-WAM~\cite{yan2026robustwam} aligns action queries with future visual features. Faster-WAM~\cite{zhao2026fasterwam} routes cached future features to selected action layers, while ForeWAM~\cite{huang2026forewam} caches future-slot features alongside dynamics registers trained with latent-action targets.
Our BFA-WAM also reuses predictive features during action generation. We supervise its future-flow slots with optical flow estimated from demonstrations. At each replanning step, a single world-model pass produces the current RGB and future-flow features used throughout action denoising.

\section{Method}
\label{sec:method}

\subsection{Problem Formulation}
\label{sec:method:setup}

We learn language-conditioned manipulation policies from demonstrations.
At each replanning step $t$, the robot has access to its current multi-view RGB observation $o_t$, a history of RGB observations $\mathcal H_t$, its proprioceptive state $s_t$, and a language instruction $\ell$.
We model the conditional distribution of the next $H$ actions as
\begin{equation}
    p_{\theta,\phi}(A_t \mid o_t,\mathcal H_t,s_t,\ell),
    \qquad A_t=(a_t,\ldots,a_{t+H-1}).
\end{equation}

Here $\theta$ and $\phi$ denote the parameters of the world and action models, respectively.

\subsection{Model Architecture}

\begin{figure*}[t]
    \centering
    \includegraphics[width=\textwidth]{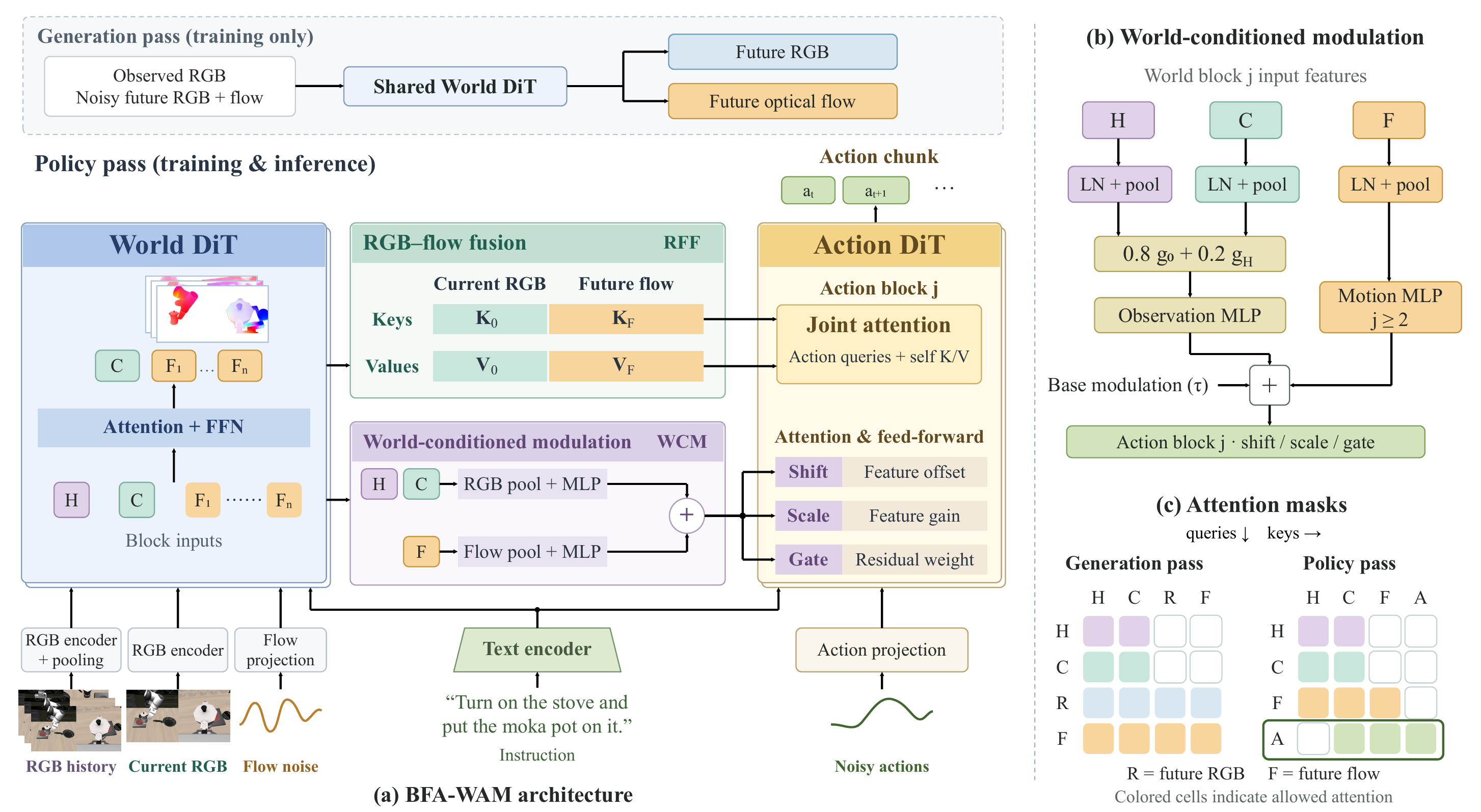}
    \caption{\textbf{Overview of BFA-WAM.}
    (a) The world diffusion transformer (World DiT) guides action denoising in the action diffusion transformer (Action DiT) through RGB--flow fusion (RFF) and world-conditioned modulation (WCM).
    (b) WCM branches with layer normalization (LN) and multilayer perceptrons (MLPs).
    (c) Attention masks for the generation and policy passes.
    Each transformer block contains attention and a feed-forward network (FFN).}
    \label{fig:architecture}
\end{figure*}

\paragraph{Overview.}
As shown in \cref{fig:architecture}(a), the world diffusion transformer (World DiT) processes the current RGB observation, a compact RGB history, and noise-initialized future-flow tokens in one forward pass at each replanning step.
RGB--flow fusion (RFF) enables the action diffusion transformer (Action DiT) to attend to keys and values from the current RGB and future-flow tokens. World-conditioned modulation (WCM) uses summaries of the observed scene and its anticipated motion to modulate the updates within the action layers.
The Action DiT reuses both the attention cache and the modulation signals as it denoises the action sequence $A_t$.

\paragraph{Current and historical RGB encoding.}
We concatenate the camera views horizontally and encode them with the frozen encoder $E$ of a video variational autoencoder (VAE).
The current observation $o_t$ retains its full patch grid.
Let $o_0$ denote the initial observation and $r_1,\ldots,r_4$ the recent observations at steps $t-8,t-6,t-4,t-2$.
The recent observations are sampled two control steps apart, whereas the gap between the episode anchor $o_0$ and $r_1$ varies with $t$.
Encoding them as one clip would present this variable gap to the VAE as a single frame transition.
We therefore encode the anchor separately and use a copy of $r_1$ as the VAE's single-frame prefix for the recent clip.
\begin{flalign}
    &\quad\begin{array}{@{}l@{\;}c@{\;}l@{}}
        z_{\rm anchor}                  & = & E(o_0),\\
        (z_{\rm prefix},z_{\rm recent}) & = & E(r_1,r_1,r_2,r_3,r_4),\\
        Z_{\mathcal H_t}                & = & [z_{\rm anchor};z_{\rm recent}].
    \end{array} &&
\end{flalign}
Here $[\,\cdot\,;\,\cdot\,]$ denotes temporal concatenation. We discard the prefix latent $z_{\rm prefix}$.
After patch projection, we apply adaptive average pooling to each historical group independently within each camera view.
In our two-camera setup, pooling each $7\times7$ view to $4\times4$ reduces the two historical groups from 196 tokens to 64, while the current observation retains all 98 tokens.
Each pooled token takes its rotary spatial coordinates from the center of its pooling region on the original grid.
The anchor and recent groups occupy the two temporal positions preceding the current group.
At the start of an episode, the anchor supplies missing recent frames, and we mask the recent group when all four frames are missing.

\subsubsection{RGB--Flow Fusion (RFF)}
\label{sec:method:rff}

RFF comprises a dual-stream RGB--flow World DiT that learns appearance and motion representations, and a future-flow conditioning module that supplies current RGB and future-flow features to the Action DiT.

\paragraph{Dual-stream RGB--flow World DiT.}
Our World expert uses a single 30-block Wan2.2 video DiT to model future appearance and motion.
RGB and optical flow each have a patch projection and an output head. We initialize the flow modules from their RGB counterparts.
For flow supervision, we compute optical flow offline with RAFT~\cite{teed2020raft} between successive sampled frames in each view.
We encode direction as hue and clipped magnitude as saturation, then pass the resulting images through the same video VAE used for RGB.
A constant first frame aligns the flow sequence with the RGB sequence. We discard its latent and retain only the two future flow groups.

The joint training pass takes four groups of visual tokens.
The first two, compressed historical RGB $X_H$ and dense current RGB $X_0$, form the clean visual condition. Both have noise level zero.
The other two represent future RGB and future flow. We perturb their latents independently at the same noise level $\sigma$ and apply the RGB and flow patch projections to obtain $X_R^\sigma$ and $X_F^\sigma$, respectively.
We concatenate all four groups along the token dimension and process the entire sequence with the same World blocks $W_\theta$.
\begin{flalign}
    &\quad\begin{array}{@{}l@{\;}c@{\;}l@{}}
        X                   & = & [X_H;X_0;X_R^\sigma;X_F^\sigma],\\
        {[U_H;U_0;U_R;U_F]} & = & W_\theta(X,\boldsymbol{\sigma};q_t),\\
        (\hat v_R,\hat v_F) & = & (D_R(U_R,\sigma),D_F(U_F,\sigma)).
    \end{array} &&
    \label{eq:world-io}
\end{flalign}
Here $\boldsymbol{\sigma}$ assigns the group noise levels $(0,0,\sigma,\sigma)$ to their respective tokens.
All groups cross-attend to $q_t$, the language embeddings of $\ell$ and a projected token for the current proprioceptive state $s_t$, which also conditions the Action DiT.
Within each World block, observed RGB queries can read only valid observed RGB keys. Future RGB and flow queries can read the observed prefix and attend to one another in both directions.
After the final block, we split the sequence at its input group boundaries. The RGB head $D_R$ and flow head $D_F$ map their respective future hidden states back to the latent grids, yielding flow-matching velocities for future RGB and future optical flow.
During action generation, the same blocks process the clean RGB prefix followed by noise-initialized future-flow tokens. Their intermediate features supply the action condition described below.

\paragraph{Future-flow conditioning of the Action DiT.}
To construct the action condition, we append two groups of noise-initialized flow tokens to the historical and current RGB tokens.
As the flow tokens attend to the observed prefix, their hidden states acquire information about the scene and its recent motion.
At each World layer $j$, we concatenate keys and values from the current RGB positions (subscript $0$) and the future-flow positions (subscript $F$) to form the policy cache
\begin{equation}
    K_P^j=[K_0^j;K_F^j], \qquad V_P^j=[V_0^j;V_F^j].
\end{equation}
Historical observations contribute to the cache through their influence on the current RGB and future-flow features.
Within the mixture of transformers (MoT), queries from layer $j$ of the action expert attend jointly to $(K_P^j,V_P^j)$ and the action tokens' own keys and values, giving each action token access to current appearance and anticipated motion.

\subsubsection{World-Conditioned Modulation (WCM)}
\label{sec:method:wcm}

RFF allows each action token to retrieve current appearance and anticipated motion through attention.
WCM complements this retrieval with a shared condition that adapts the feature transformations and residual updates within each Action block.
It derives observation and motion summaries from the same World pass, using separate branches to map them to layer-wise modulation residuals.

At layer $j$, we normalize the hidden states entering the World block and average the current RGB, valid historical RGB, and future-flow tokens separately, obtaining $g_0^j$, $g_H^j$, and $g_F^j$.
Two layer-specific multilayer perceptrons (MLPs) produce the observation and motion residuals:
\begin{equation}
    \begin{aligned}
        \Delta m_{\rm obs}^j &= M_{\rm obs}^j(0.8g_0^j+0.2g_H^j),\\
        \Delta m_F^j &= M_F^j(g_F^j).
    \end{aligned}
    \label{eq:wcm-branches}
\end{equation}
The observation branch emphasizes the current scene while retaining temporal context from the RGB history.
Pooling the two groups separately makes their relative contribution independent of their token counts.
Invalid historical tokens are excluded from the average, and we use $g_0^j$ alone when no historical token is valid.

The motion branch summarizes the future-flow features to condition action updates on anticipated scene motion.
We enable this branch from layer 2 onward: the flow tokens entering that layer have already attended to the observed prefix in the first World block.
This ensures that motion modulation uses features that have incorporated observation information.

For action noise level $\tau$, we add both residuals to the Action block's base modulation:
\begin{equation}
    m_A^j(\tau)=m_{\rm base}^j(\tau)+\Delta m_{\rm obs}^j
    +\mathbf{1}_{j\geq 2}\,\Delta m_F^j .
\end{equation}
Here $m_{\rm base}^j(\tau)$ combines the projected noise-level embedding with the block's learned modulation parameters.
Each residual supplies six channel-wise vectors: a shift, scale, and gate for each of the attention and feed-forward sublayers.
The shifts and scales adjust normalized action features before each sublayer, while the gates weight its contribution to the residual update.
These vectors are shared across action positions, providing a common scene and motion condition for the entire action chunk.

We zero-initialize the final projections of both branches, so WCM initially preserves the base modulation.
At inference, the world-conditioned residuals are computed once per replanning step and reused throughout action denoising; the base modulation continues to vary with $\tau$.

\subsubsection{Training and Inference}
We optimize the World and Action DiTs jointly by minimizing
\begin{equation}
    \begin{aligned}
        \mathcal{L} ={}&
        \lambda_{\rm rgb}\mathcal{L}_{\rm rgb}^{\rm FM}
        + \lambda_{\rm flow}\mathcal{L}_{\rm flow}^{\rm FM} \\
        &+ \lambda_{\rm act}\mathcal{L}_{\rm act}^{\rm FM}
        + \lambda_{\rm 1step}\mathcal{L}_{\rm flow}^{\rm 1step}.
    \end{aligned}
    \label{eq:training-objective}
\end{equation}
The three flow-matching (FM) terms regress the velocity $\epsilon-x$ for a clean target $x$ and Gaussian noise $\epsilon$~\cite{lipman2022flowmatching}. The four coefficients default to one.

The first World pass (\cref{eq:world-io}) computes $\mathcal{L}_{\rm rgb}^{\rm FM}$ and $\mathcal{L}_{\rm flow}^{\rm FM}$ on the future RGB and optical-flow latents. At noise levels $\sigma<1$, the inputs retain information about the demonstrated future that is unavailable to the policy at deployment.

We therefore run the same World DiT a second time with the clean observation prefix and future-flow slots initialized from fresh Gaussian noise $\epsilon_F$ at $\sigma=1$.
The Action DiT uses the resulting K/V and modulation to denoise an independently perturbed action sequence under $\mathcal{L}_{\rm act}^{\rm FM}$.
From this same World pass, the flow head predicts a velocity $\hat v_F$. The one-step loss $\mathcal{L}_{\rm flow}^{\rm 1step}$ is the mean squared error between the reconstructed latent $\epsilon_F-\hat v_F$ and the encoded demonstration flow, averaged over valid future-flow groups.
This term supervises the motion forecast under the input conditions used for action generation.

At inference, only the second World pass is needed at each replanning step.

\section{Experiments}
\label{sec:experiments}

We evaluate BFA-WAM on LIBERO~\cite{liu2023libero}, LIBERO-Plus~\cite{fei2025liberoplus}, and RoboTwin 2.0~\cite{chen2025robotwin2} to study manipulation performance, robustness to distribution shifts, and the contributions of RFF and WCM.

\subsection{Experimental Setup}
\label{sec:experiments:setup}

We use task success rate as the primary metric across the three benchmarks.

\paragraph{LIBERO.}
LIBERO~\cite{liu2023libero} evaluates language-conditioned manipulation through the Spatial, Object, Goal, and Long suites, covering spatial relations, object variation, goal specification, and long-horizon tasks.
Each suite contains ten tasks.
For BFA-WAM, we jointly train a single policy across all four suites for 20,000 steps with an effective batch size of 128, after removing no-op actions from the demonstrations.
The policy takes external and wrist-camera images, language instructions, and proprioception as input.
We run 50 trials per task and report the success rate of each suite and the mean across the four suites.

\paragraph{RoboTwin 2.0.}
RoboTwin 2.0~\cite{chen2025robotwin2} evaluates coordinated bimanual manipulation across diverse scene conditions.
We use 50 tasks and jointly train a single policy for five epochs on 2,500 clean-scene and 25,000 randomized-scene demonstrations.
The policy receives images from a head camera and two wrist cameras, together with language instructions and proprioception, and predicts 14-dimensional actions.
Each task is evaluated over 100 trials in both Clean and Randomized settings using held-out instructions.
The Randomized setting varies backgrounds, clutter, lighting, and table height.
We report task-averaged success rates for both settings and their mean.

\paragraph{LIBERO-Plus.}
LIBERO-Plus~\cite{fei2025liberoplus} tests robustness to seven types of distribution shift involving camera viewpoints, robot initial states, language instructions, lighting, background textures, sensor noise, and object layouts.
We use the policies trained on LIBERO directly, retaining their training-set normalization statistics.
We evaluate all 10,030 task variants with one rollout per variant.
Category success rates pool rollouts within each category; overall success pools all evaluated rollouts across the four suites, dividing total successes by total rollouts.

\paragraph{Implementation details.}
We initialize the World DiT from Wan2.2-TI2V-5B, using 30 blocks with a hidden dimension of 3,072.
Continuous actions are modeled by a 30-block Action DiT with a hidden dimension of 1,024.
We freeze the VAE and text encoder, cache the text embeddings, and jointly optimize both DiTs, the proprioceptive projection, and the added modules.
Training runs on eight NVIDIA H20 GPUs in bfloat16, with four samples per GPU and four gradient accumulation steps.
We use AdamW with a learning rate of $10^{-4}$, weight decay of $10^{-2}$, and momentum coefficients $(0.9,0.95)$.
The first 5\% of training uses linear warmup, followed by cosine decay to $10^{-6}$, with the gradient norm clipped at 1.0.
All four loss coefficients in \cref{eq:training-objective} are set to one in the full model.
At inference, the policy generates 32-action chunks using ten denoising steps and a text guidance scale of 1.0.
We execute the first ten actions on LIBERO and LIBERO-Plus, and the first 24 on RoboTwin, before replanning.

\begin{table}[t]
    \centering
    \caption{Success rates (\%) on LIBERO. P.T. denotes embodied pretraining. Bold and underline mark the highest and second-highest scores in each column, respectively.}
    \label{tab:libero-main}
    \setlength{\tabcolsep}{2.6pt}
    \footnotesize
    \begin{tabular}{l|c|rrrrr}
        \toprule
        Method & P.T. & Spatial & Object & Goal & Long & Avg. \\
        \midrule
        OpenVLA~\cite{kim2024openvla} & \ding{51} & 84.7 & 88.4 & 79.2 & 53.7 & 76.5 \\
        WorldVLA~\cite{cen2025worldvla} & \ding{51} & 87.6 & 96.2 & 83.4 & 60.0 & 81.8 \\
        $\pi_0$-FAST~\cite{pertsch2025fast} & \ding{51} & 96.4 & 96.8 & 88.6 & 60.2 & 85.5 \\
        $\pi_0$~\cite{black2025pi0} & \ding{51} & 96.8 & 98.8 & 95.8 & 85.2 & 94.1 \\
        UniVLA~\cite{bu2025univla} & \ding{51} & 96.5 & 96.8 & 95.6 & 92.0 & 95.2 \\
        $\pi_{0.5}$~\cite{black2025pi05} & \ding{51} & 98.8 & 98.2 & \underline{98.0} & 92.4 & 96.9 \\
        CogVLA~\cite{li2025cogvla} & \ding{51} & 98.6 & 98.8 & 96.6 & 95.4 & 97.4 \\
        Motus~\cite{bi2025motus} & \ding{51} & 96.8 & \underline{99.8} & 96.6 & 97.6 & 97.7 \\
        LingBot-VA~\cite{li2026lingbotva} & \ding{51} & 98.5 & 99.6 & 97.2 & \textbf{98.5} & \underline{98.5} \\
        Fast-WAM~\cite{yuan2026fastwam} & \ding{55} & 98.2 & \textbf{100.0} & 97.0 & 95.2 & 97.6 \\
        Fast-WAM-Joint~\cite{yuan2026fastwam} & \ding{55} & \textbf{99.6} & 99.4 & \textbf{98.2} & 96.8 & \underline{98.5} \\
        ForeWAM~\cite{huang2026forewam} & \ding{55} & 97.6 & \underline{99.8} & \underline{98.0} & 96.2 & 97.9 \\
        \midrule
        BFA-WAM (ours) & \ding{55} & \underline{99.4} & \underline{99.8} & \textbf{98.2} & \underline{98.4} & \textbf{99.0} \\
        \bottomrule
    \end{tabular}
\end{table}

\begin{table}[t]
    \centering
    \caption{Success rates (\%) on RoboTwin 2.0 under clean and randomized conditions. P.T. denotes embodied pretraining. Bold and underline mark the highest and second-highest scores in each column, respectively.}
    \label{tab:robotwin-main}
    \setlength{\tabcolsep}{5pt}
    \footnotesize
    \begin{tabular}{l|c|rrr}
        \toprule
        Method & P.T. & Clean & Rand. & Avg. \\
        \midrule
        $\pi_0$~\cite{black2025pi0} & \ding{51} & 65.9 & 58.4 & 62.2 \\
        $\pi_{0.5}$~\cite{black2025pi05} & \ding{51} & 82.7 & 76.8 & 79.8 \\
        Motus~\cite{bi2025motus} & \ding{51} & 88.7 & 87.0 & 87.8 \\
        Motus (Wan2.2 init.)~\cite{bi2025motus,yuan2026fastwam} & \ding{55} & 77.6 & 77.0 & 77.3 \\
        LingBot-VA~\cite{li2026lingbotva} & \ding{51} & \underline{92.9} & 91.5 & 92.2 \\
        Fast-WAM~\cite{yuan2026fastwam} & \ding{55} & 91.9 & 91.8 & 91.8 \\
        Fast-WAM-Joint~\cite{yuan2026fastwam} & \ding{55} & 90.8 & 90.3 & 90.6 \\
        FlowWAM~\cite{chen2026flowwam} & \ding{51} & \underline{92.9} & 92.1 & 92.5 \\
        Faster-WAM~\cite{zhao2026fasterwam} & \ding{55} & 92.8 & \underline{92.3} & \underline{92.6} \\
        \midrule
        BFA-WAM (ours) & \ding{55} & \textbf{93.4} & \textbf{93.6} & \textbf{93.5} \\
        \bottomrule
    \end{tabular}
\end{table}

\begin{table*}[t]
    \centering
    \caption{Success rates (\%) on LIBERO-Plus across seven distribution shifts. Overall pools all evaluation rollouts. P.T. denotes embodied pretraining. Bold and underline mark the highest and second-highest scores in each column, respectively.}
    \label{tab:plus-main}
    \setlength{\tabcolsep}{5pt}
    \footnotesize
    \begin{tabular}{l|c|rrrrrrrr}
        \toprule
        Method & P.T. & Cam. & Robot & Lang. & Light & Back. & Noise & Layout & Overall \\
        \midrule
        OpenVLA~\cite{kim2024openvla} & \ding{51} & 0.8 & 3.5 & 23.0 & 8.1 & 34.8 & 15.2 & 28.5 & 15.6 \\
        OpenVLA-OFT~\cite{kim2025openvlaoft} & \ding{51} & 56.4 & 31.9 & 79.5 & 88.7 & \textbf{93.3} & 75.8 & 74.2 & 69.6 \\
        $\pi_0$~\cite{black2025pi0} & \ding{51} & 13.8 & 6.0 & 58.8 & 85.0 & 81.4 & \underline{79.0} & 68.9 & 53.6 \\
        $\pi_0$-FAST~\cite{pertsch2025fast} & \ding{51} & \textbf{65.1} & 21.6 & 61.0 & 73.2 & 73.2 & 74.4 & 68.8 & 61.6 \\
        UniVLA~\cite{bu2025univla} & \ding{51} & 1.8 & 46.2 & 69.6 & 69.0 & 81.0 & 21.2 & 31.9 & 42.9 \\
        WorldVLA~\cite{cen2025worldvla} & \ding{51} & 0.1 & 27.9 & 41.6 & 43.7 & 17.1 & 10.9 & 38.0 & 25.0 \\
        Fast-WAM~\cite{yuan2026fastwam} & \ding{55} & 19.6 & 46.0 & 71.3 & 81.5 & 47.6 & 31.1 & 60.0 & 49.5 \\
        ForeWAM~\cite{huang2026forewam} & \ding{55} & \underline{61.7} & 56.8 & 85.1 & 95.8 & 76.5 & \textbf{89.1} & \textbf{83.0} & \underline{77.6} \\
        Faster-WAM~\cite{zhao2026fasterwam} & \ding{55} & 53.8 & \underline{71.6} & \textbf{94.7} & \underline{96.3} & 61.3 & 63.6 & 79.1 & 73.6 \\
        \midrule
        BFA-WAM (ours) & \ding{55} & 60.4 & \textbf{80.8} & \underline{93.4} & \textbf{96.9} & \underline{82.7} & 78.2 & \underline{80.5} & \textbf{81.1} \\
        \bottomrule
    \end{tabular}
\end{table*}

\subsection{Main Results}
\label{sec:experiments:main}

\paragraph{LIBERO.}
\Cref{tab:libero-main} summarizes success rates across the four LIBERO suites.
BFA-WAM achieves 99.0\% average success without embodied pretraining and reaches at least 98.2\% on every suite.
Several policies already approach the benchmark ceiling. Fast-WAM and ForeWAM average 97.6\% and 97.9\%, respectively, while Fast-WAM-Joint and the pretrained LingBot-VA both reach 98.5\%.
BFA-WAM outperforms these baselines on average.

\paragraph{RoboTwin 2.0.}
\Cref{tab:robotwin-main} compares success rates in clean and randomized scenes across the 50 bimanual tasks in RoboTwin 2.0.
BFA-WAM leads in both settings, with success rates of 93.4\% and 93.6\%, respectively.
Among policies trained without embodied pretraining, its 93.5\% average exceeds those of Faster-WAM (92.6\%), Fast-WAM (91.8\%), and Fast-WAM-Joint (90.6\%).
It also outperforms the pretrained FlowWAM and LingBot-VA in both settings. 

\paragraph{LIBERO-Plus.}
\Cref{tab:plus-main} reports success rates across the seven distribution shifts in LIBERO-Plus.
Fast-WAM and BFA-WAM both exceed 97.0\% success on standard LIBERO, but the gap between them widens under these shifts.
BFA-WAM retains an overall success rate of 81.1\%, ahead of Faster-WAM (73.6\%), OpenVLA-OFT (69.6\%), and Fast-WAM (49.5\%).
BFA-WAM outperforms Fast-WAM in all seven categories and each of the other two policies in six.
The largest gain over Faster-WAM is on background shifts (82.7\% vs.\ 61.3\%); that over OpenVLA-OFT is on robot initial-state shifts (80.8\% vs.\ 31.9\%).

\subsection{Ablation Studies}
\label{sec:experiments:ablation}

We train each variant on LIBERO with the same data and optimization schedule and evaluate it on the same LIBERO-Plus tasks and seeds. Overall success pools all evaluation rollouts.

\paragraph{Component contributions.}
\Cref{tab:module-ablation} compares two ways of conditioning the action expert on world features. The baseline attends to current RGB features and retains the future RGB and action training objectives, reaching 49.5\% overall success.
RFF adds a supervised future-flow stream whose features the action expert can attend to alongside current RGB. Even without RGB history or modulation, this variant raises overall success to 76.8\%. Its largest gains occur on Long (42.3\% to 77.3\%) and Spatial (50.4\% to 83.9\%).
WCM-only retains the full model's RGB history, future-flow stream, all four training losses, and both WCM branches, but removes future-flow K/V from action attention while retaining current RGB K/V. Future-motion features thus condition the action layers only through WCM, with motion modulation active from the second layer onward. This variant improves success over the baseline in all four suites, from 49.5\% to 60.2\% overall, with the largest increase on Goal (37.5\% to 52.8\%).

The full model makes future-flow features available through both attention and WCM. Restoring future-flow K/V to WCM-only raises overall success from 60.2\% to 81.1\%, showing the benefit of direct attention to motion features alongside modulation. Relative to RFF alone, the largest gains occur on Object (81.2\% to 92.1\%) and Goal (64.8\% to 74.3\%).

\begin{table}[!htbp]
    \centering
    \caption{Component ablation on LIBERO-Plus across the four task suites. WCM-only disables future-flow K/V in action attention while retaining both WCM branches. All values are success rates (\%).}
    \label{tab:module-ablation}
    \setlength{\tabcolsep}{3pt}
    \footnotesize
    \begin{tabular}{lrrrrr}
        \toprule
        Variant & Spatial & Object & Goal & Long & Overall \\
        \midrule
        Baseline & 50.4 & 67.7 & 37.5 & 42.3 & 49.5 \\
        RFF-only & 83.9 & 81.2 & 64.8 & 77.3 & 76.8 \\
        WCM-only & 63.0 & 77.7 & 52.8 & 47.3 & 60.2 \\
        Full & 81.2 & 92.1 & 74.3 & 77.0 & 81.1 \\
        \bottomrule
    \end{tabular}
\end{table}

\begin{figure*}[!t]
    \centering
    \includegraphics[width=\textwidth]{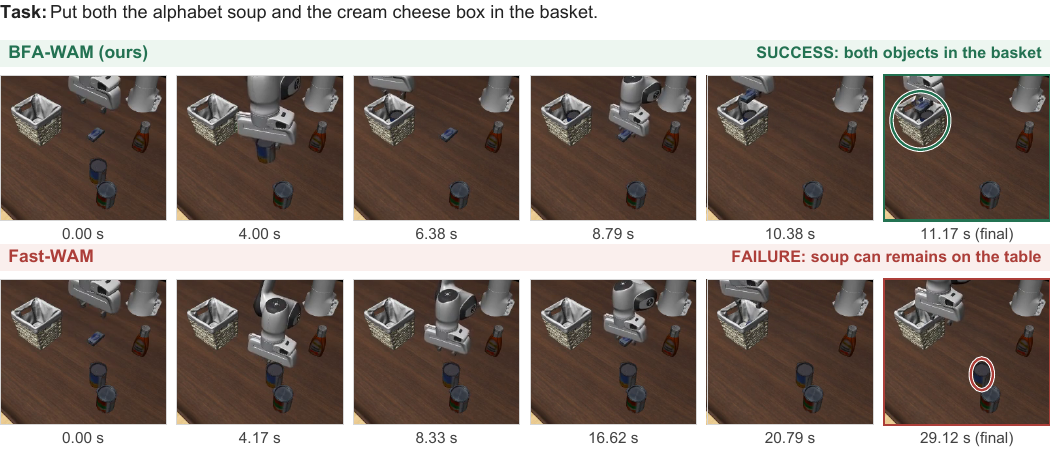}
    \caption{\textbf{Qualitative rollout comparison.} External-camera keyframes for the task ``put both the alphabet soup and the cream cheese box in the basket.'' BFA-WAM places both target objects in the basket (top), whereas Fast-WAM leaves the soup can on the table (bottom).  The final-frame outlines highlight the completed goal and the unplaced soup can.}
    \label{fig:rollout-comparison}
\end{figure*}

\paragraph{Observation history.}
\Cref{tab:design-ablation} varies one setting at a time from the full model.
We first vary the weight $\alpha$ assigned to current RGB features in the WCM observation summary (\cref{eq:wcm-branches}).
With $\alpha=1$, the summary uses only current RGB features, although the World DiT still processes historical tokens. Assigning 20\% of the summary weight to history raises success from 80.0\% to 81.1\%. Increasing the history weight further to 50\% lowers success to 78.2\%.
Among the three settings, giving the current scene more weight than its history works best for conditioning action refinement.

\paragraph{Historical token compression.}
\Cref{tab:design-ablation} also compares the spatial resolution of the history at the default weighting ($\alpha=0.8$).
Both variants use the episode's initial frame and the same four recent frames. Pooling each historical view from $7\times7$ to $4\times4$ reduces the two-camera history from 196 tokens to 64, while the current observation retains its full 98 tokens.
The pooled representation reaches 81.1\% success, compared with 78.2\% for the full historical grid. A coarser spatial summary of the same past observations thus gives higher success with fewer history tokens.

\begin{table}[!htbp]
    \centering
    \caption{Design ablations on LIBERO-Plus. We report overall success rates (\%) for the full model and variants that change history inputs, the weighting of current and historical RGB features, history pooling, one-step flow supervision, or future-motion modulation.}
    \label{tab:design-ablation}
    \setlength{\tabcolsep}{5pt}
    \footnotesize
    \begin{tabular}{lr}
        \toprule
        Variant & Overall (\%) \\
        \midrule
        Full & 81.1 \\
        w/o history & 78.4 \\
        Equal RGB weights ($\alpha=0.5$) & 78.2 \\
        Current-only pooling ($\alpha=1.0$) & 80.0 \\
        Uncompressed history ($7\times7$) & 78.2 \\
        w/o one-step loss & 69.6 \\
        w/o future modulation & 77.6 \\
        \bottomrule
    \end{tabular}
\end{table}

\paragraph{One-step flow supervision.}
\Cref{tab:design-ablation} shows the largest drop when we set the one-step flow reconstruction weight in \cref{eq:training-objective} to zero.
This variant still trains on future flow through flow matching and supplies future-flow features to both RFF and WCM. During flow matching, partially noised inputs can retain information from the demonstrated future. During action generation, all future-flow tokens start from noise.
The one-step loss supervises reconstruction of the demonstrated flow from this latter input. Removing it lowers overall success from 81.1\% to 69.6\%, an 11.5-point drop that shows the value of supervising motion under the conditions used by the policy.

\paragraph{Future-motion modulation.}
\Cref{tab:design-ablation} also tests how the action expert uses the predicted motion.
We remove the motion residual $\Delta m_F^j$ while retaining observation modulation, future-flow K/V, and all four training losses.
Each action token can still retrieve motion cues through attention. In the full model, WCM also pools these features into a shared signal that adjusts the shifts, scales, and gates of the action layers across all action positions.
Removing this shared signal lowers success from 81.1\% to 77.6\%. The 3.5-point difference measures the benefit of using the motion forecast to modulate action refinement when the same forecast is already available through attention.

\subsection{Qualitative Analysis}
\label{sec:experiments:qualitative}

As shown in \cref{fig:rollout-comparison}, we compare BFA-WAM and Fast-WAM on placing the alphabet soup can and cream cheese box in a basket.
This task requires selecting both targets among distractors and completing successive grasp-and-place actions.
BFA-WAM first transfers the soup can and then the cream cheese box, completing the instruction.
Fast-WAM approaches the soup can without transferring it, then manipulates the cream cheese box while leaving the can on the table.

\section{Conclusion}
\label{sec:conclusion}

In this paper, we presented BFA-WAM, a world action framework that connects explicit future-motion supervision with world-conditioned action refinement.
The world model jointly learns future RGB and optical-flow prediction, while one-step flow reconstruction supervises its motion features under the input conditions used at inference.
These world features guide action generation through two complementary mechanisms.
WCM summarizes observed RGB and future-flow features through separate branches, producing layer-wise shifts, scales, and gates shared across action positions.
RFF enables action queries to attend jointly to current RGB and flow-supervised future features, combining visible scene context with anticipated motion.
At inference, one world-model pass supplies both signals for all denoising steps without decoding future frames or estimating optical flow online.
Experiments across multiple robotic manipulation benchmarks demonstrate the effectiveness of BFA-WAM.
The benchmarks span single-arm and bimanual tasks.
BFA-WAM also remains effective under distribution shifts.
These results support using latent motion forecasts to guide closed-loop manipulation.

\twocolumn[{%
  \centering
  \pdfbookmark[0]{Supplementary Material}{supplementary}%
  \Large\vphantom{\textbf{\thetitle}}\\
  \vspace{.5em}\textbf{Supplementary Material}\vphantom{p}\\
  \vspace{1em}%
}]
\appendix
\suppressfloats[t]
\begin{figure*}[t]
\centering
\includegraphics[width=\textwidth]{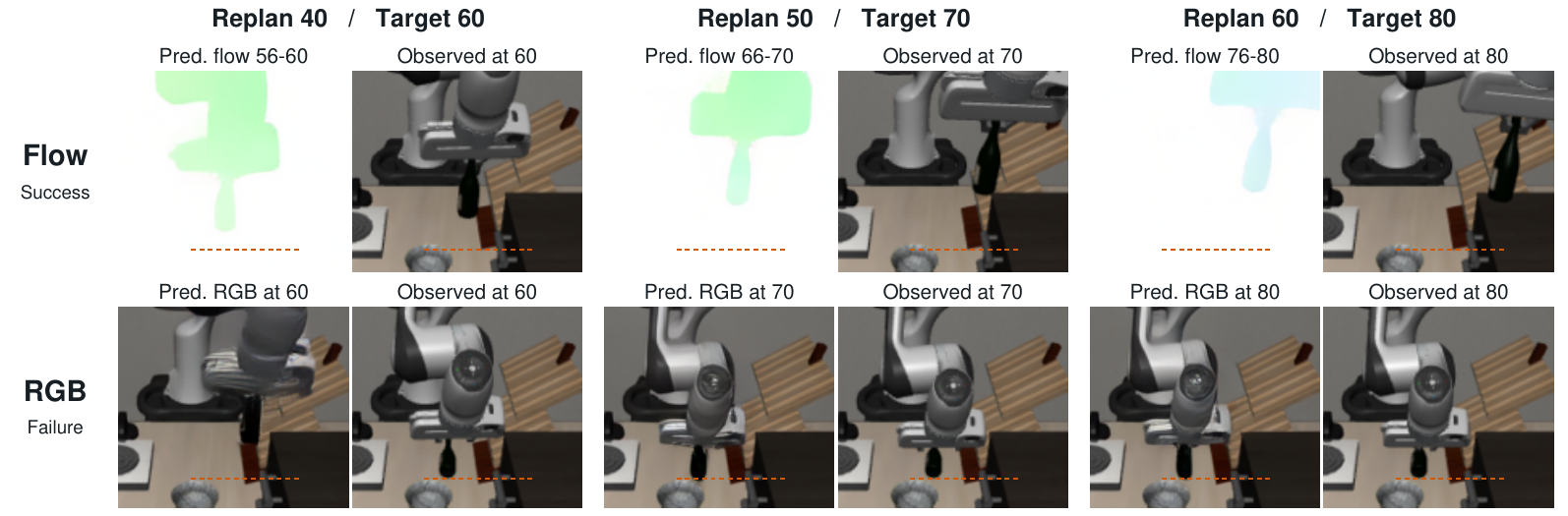}
\caption{Predicted and executed bottle lifts across three replans. Each pair compares a forecast (left) with the observed camera frame 20 control steps later (right). The flow panels show predicted motion over the final four steps of each forecast. Dashed lines mark the bottle base at step 40 in each rollout. At the first replan, both policies anticipate a lift, but only the flow-conditioned policy lifts the bottle by step 60.}
\label{fig:supp-future}
\end{figure*}

\section{Training Details of BFA-WAM}
\label{sec:supp-implementation}
The main paper describes the architecture and learning objectives of BFA-WAM. Here, we provide the optimization settings and data construction used for training on LIBERO~\cite{liu2023libero} and RoboTwin 2.0~\cite{chen2025robotwin2}. We jointly optimize the World DiT, Action DiT, proprioceptive projection, and the added RGB--flow fusion and world-conditioned modulation modules. The video VAE and text encoder remain frozen, and task instructions are provided through cached language embeddings. LIBERO-Plus~\cite{fei2025liberoplus} uses the LIBERO-trained policy without additional training.

Training uses bfloat16 precision and DeepSpeed ZeRO-1. We use AdamW with momentum coefficients $(0.9,0.95)$, a learning rate of $10^{-4}$, and weight decay of $10^{-2}$. A linear warm-up over the first 5\% of training is followed by cosine decay to $10^{-6}$, with the gradient norm clipped to 1.0. The World and Action flow schedulers each use 1,000 training timesteps, with shifts of 5.0 and 1.0, respectively. The four objectives---RGB flow matching, optical-flow flow matching, action flow matching, and one-step flow reconstruction---have equal weights. Flow supervision excludes invalid future targets. Table~\ref{tab:supp-settings} summarizes the common optimization settings.

\begin{table}[t]
\centering
\caption{Common training hyperparameters for BFA-WAM.}
\label{tab:supp-settings}
\small
\setlength{\tabcolsep}{4pt}
\begin{tabular}{@{}ll@{}}
\toprule
Parameter & Value \\
\midrule
Distributed strategy & DeepSpeed ZeRO-1 \\
Precision & bfloat16 \\
Optimizer & AdamW, $\beta=(0.9,0.95)$ \\
Learning rate & $10^{-4}$ \\
Weight decay & $10^{-2}$ \\
LR schedule & 5\% warm-up, cosine to $10^{-6}$ \\
Gradient clipping & 1.0 \\
Flow training timesteps & 1,000 (World and Action) \\
Scheduler shift & 5.0 (World), 1.0 (Action) \\
Loss weights & All four coefficients set to 1.0 \\
Training seed & 42 \\
\bottomrule
\end{tabular}
\end{table}

Each training sample covers 33 consecutive observation--state pairs and 32 action steps. Sampling visual observations at offsets $[0,4,\ldots,32]$ gives one current and eight future frames. For LIBERO, the external and wrist-camera views are resized to $224\times224$ and concatenated horizontally into a $224\times448$ input. We train one policy across the four LIBERO suites for 10 epochs after removing no-op actions, using training-set min/max statistics to normalize actions and states. For RoboTwin, the head view and two wrist views form a composite $384\times320$ input; actions and states use z-score normalization. The RoboTwin training set contains 2,500 clean-scene and 25,000 randomized-scene demonstrations. We use eight NVIDIA H20 GPUs for LIBERO and 64 for RoboTwin, with four samples per GPU and four gradient accumulation steps, giving effective batch sizes of 128 and 1,024, respectively. Dataset-specific configurations are summarized in Table~\ref{tab:supp-data-settings}.

\begin{table}[t]
\centering
\caption{Dataset-specific training configurations. The visual sequence includes one current and eight future observations, in addition to the RGB history.}
\label{tab:supp-data-settings}
\small
\setlength{\tabcolsep}{4pt}
\begin{tabular}{@{}lcc@{}}
\toprule
Setting & LIBERO & RoboTwin 2.0 \\
\midrule
GPUs & 8 & 64 \\
Input views & 2 & 3 \\
Resolution ($H\times W$) & $224\times448$ & $384\times320$ \\
Visual sequence & $1+8$ & $1+8$ \\
Visual stride & 4 & 4 \\
Action chunk & 32 & 32 \\
Action dimension & 7 & 14 \\
Micro-batch/GPU & 4 & 4 \\
Gradient accumulation & 4 & 4 \\
Global batch & 128 & 1,024 \\
Training duration & 10 epochs & 5 epochs \\
\bottomrule
\end{tabular}
\end{table}

The historical RGB input contains the episode anchor and recent observations at $t-8$, $t-6$, $t-4$, and $t-2$. We encode the anchor separately and prepend a duplicate of the first recent frame when encoding the recent clip, discarding the resulting prefix latent. We compress historical tokens on both LIBERO and RoboTwin by independently pooling each camera region to a $4\times4$ grid within each historical latent group. Spatial rotary coordinates retain the pooling-region centers in the original composite grid. Current-observation tokens remain uncompressed: 98 for LIBERO and 120 for RoboTwin. Historical features contribute through World attention and WCM, but historical keys and values are not directly included in the action-attention cache. The observation modulation combines current and historical summaries with weights 0.8 and 0.2; the observation and motion modulation MLPs have hidden dimensions 1,024 and 256, respectively.

Future-flow targets are computed offline with RAFT~\cite{teed2020raft} between successive sampled frames, covering $t\!\rightarrow\!t+4$ through $t+28\!\rightarrow\!t+32$. Direction is encoded as hue and magnitude as saturation, with a configured clipping bound of 64 pixels. A constant white frame prefixes the flow sequence for VAE encoding; its latent is discarded, leaving two future-flow latent groups for supervision. Training uses two World forward passes: the first learns joint future RGB and flow generation, and the second constructs action conditions from clean observations and noise-initialized future-flow slots while also providing one-step reconstruction supervision. This second pass matches the conditioning procedure at deployment, where its keys, values, and modulation residuals are computed once and reused throughout action denoising.

\section{Future RGB versus Future Flow}
\label{sec:supp-ablation}
Given the same RGB observations, which representation of the future best guides the robot's next actions? We compare two ways to condition the Action DiT: pairing current RGB features with predicted future RGB features, or pairing them with predicted future optical-flow features. In the latter pairing, current RGB identifies the gripper, the object, and the goal in the scene, while future flow describes how image regions are expected to move between observations.

\subsection{Policy Performance on LIBERO-Plus}
We vary the modality of the future features supplied to the policy while retaining RGB history, the two World DiT training passes, and ten action denoising steps. During training, the first pass learns future prediction from perturbed targets. The second pass constructs the policy condition from clean observations and noise-initialized future slots. At inference, both variants reuse the intermediate features of a single World DiT pass throughout action denoising.

\begin{table}[t]
\centering
\caption{Action conditioning on LIBERO-Plus. The two policies share the same RGB observation inputs and action denoising schedule.}
\label{tab:supp-ablation}
\small
\begin{tabular}{@{}lc@{}}
\toprule
Features for action conditioning & Success rate (\%) $\uparrow$ \\
\midrule
Current RGB + future RGB & 72.1 \\
Current RGB + future flow (BFA-WAM) & \textbf{81.1} \\
\bottomrule
\end{tabular}
\end{table}

Pairing current RGB with future flow raises task success from 72.1\% to 81.1\% on LIBERO-Plus (Table~\ref{tab:supp-ablation}). The 9.0-point gain favors future flow as the intermediate modality for action conditioning in this setting. One possible explanation is that flow makes the direction and extent of anticipated motion explicit, while current RGB retains the appearance cues needed to locate that motion in the scene.

\subsection{Visual Comparison}
Figure~\ref{fig:supp-future} illustrates the distinction on the LIBERO-Goal bottle-placement task. At step 40, both policies anticipate a lift: the RGB forecast depicts the bottle above its initial position, and the flow forecast localizes motion to the gripper and bottle. Yet their observations at step 60 differ. The bottle remains at the dashed line in the RGB rollout and rises above it in the flow rollout. By step 80, the flow-conditioned policy has carried the bottle over the cabinet. It completes the placement at step 87, while the RGB-conditioned policy fails within 400 steps. Although both policies predicted a lift, only the flow-conditioned action expert generated actions that lifted the bottle. This suggests that future flow provides more effective guidance than future RGB for generating the lifting motion in this task.

\section{Inference Efficiency}
\label{sec:supp-efficiency}
To balance inference latency and task success, we introduce BFA-WAM-Fast, which reduces action denoising from ten steps to two. Table~\ref{tab:supp-latency} compares the inference latency of Fast-WAM~\cite{yuan2026fastwam}, BFA-WAM, and BFA-WAM-Fast on NVIDIA H20. Each latency is a mean over timed policy calls after warm-up.

\noindent\begin{minipage}{\columnwidth}
\centering
\captionof{table}{Inference latency on NVIDIA H20 (ms per policy call) and LIBERO-Plus success rate (SR). Total latency is measured over the full policy call. A dash indicates that the stage had no recorded calls.}
\label{tab:supp-latency}
\small
\setlength{\tabcolsep}{2.5pt}
\begin{tabular}{@{}lrrr@{}}
\toprule
 & Fast-WAM & BFA-WAM & \shortstack{BFA-WAM\\-Fast} \\
\midrule
Current-frame VAE & 10.9 & 15.3 & 11.5 \\
History VAE & --- & 50.6 & 47.9 \\
Context & 18.4 & 26.2 & 20.1 \\
World prefill & 25.3 & 64.4 & 52.3 \\
Action denoising & 244.5 & 345.1 & 56.1 \\
\midrule
Total latency & 299.3 & 524.9 & \textbf{194.3} \\
SR (\%) & 49.5 & \textbf{81.1} & 79.3 \\
\bottomrule
\end{tabular}
\end{minipage}\par\medskip

Compared with the original Fast-WAM, BFA-WAM-Fast reduces mean policy latency from 299.3 to 194.3\,ms per call on NVIDIA H20, a $1.54\times$ inference speedup. On LIBERO-Plus, task success increases from 49.5\% to 79.3\%, a gain of 29.8 percentage points. BFA-WAM prioritizes task success, while BFA-WAM-Fast balances inference speed and task success.

{\small\raggedbottom
\bibliographystyle{ieeenat_fullname}
\pdfbookmark[0]{\refname}{references}
\bibliography{main}
}

\end{document}